%% file: main.tex
\documentclass[letterpaper,10pt,conference]{ieeeconf}

\IEEEoverridecommandlockouts
\usepackage[utf8]{inputenc}
\usepackage[T1]{fontenc}
\usepackage{url}
\usepackage{graphicx}
\usepackage{float}
\title{\LARGE \bf
Learning Expressive Humanoid Locomotion from Monocular Runway Videos for Robot Fashion Shows
}

\author{Kyrylo Kolesnichenko$^{1,2}$, Irvin Steve Cardenas$^{2}$, and Jong-Hoon Kim$^{2}$%
\thanks{$^{1}$Kaunas Faculty, Vilnius University, Lithuania}%
\thanks{$^{2}$Advanced Telerobotics Research Lab - Computer Science Department, Kent State University, Kent, OH 44242, USA
        {\tt\small jkim72@kent.edu}}%
}

\begin{document}

\maketitle
\thispagestyle{empty}
\pagestyle{empty}

\input{sections/abstract}

\input{sections/introduction}

\input{sections/related_work}

\input{sections/method}

\input{sections/experimental_setup}

\input{sections/results}

\input{sections/future_work}

\input{main.bbl}
\end{document}

%% file: sections/abstract.tex
\begin{abstract}
Runway walking requires coordinated control of posture, stride, foot placement, and whole-body motion to effectively present clothing and convey a distinctive style. However, humanoid robots used in fashion shows typically rely on locomotion policies optimized primarily for stability and walking speed, limiting their ability to reproduce expressive, human-like runway motions. In this work, we present an end-to-end framework that transforms monocular runway videos into deployable humanoid locomotion policies through motion recovery, robot retargeting, motion correction, policy training, simulation-based evaluation, and physical deployment. We evaluate the proposed framework on the Booster K1 humanoid robot using runway-style catwalk motions. The learned policy completed every physical trial without falling, while reproducing the characteristic narrow foot placement and coordinated movement of the legs, torso, and arms. The results demonstrate that our proposed training framework enables the Booster K1 to perform stable and expressive catwalk motions, highlighting its potential for humanoid robotic applications in fashion shows and other performance-oriented scenarios.
\end{abstract}

%% file: sections/introduction.tex
\section{INTRODUCTION}

The catwalk is a specialized form of expressive locomotion in which posture, cadence, stride, foot placement, arm motion, gaze, posing, and turning are coordinated to present a garment and communicate the aesthetic of a fashion show. Unlike ordinary walking, its objective is not only stable movement, but also the control of silhouette, drape, rhythm, and perceived character. Catwalk styles vary across designers, garments, footwear, and show direction, making runway walking a context-dependent performance rather than a single fixed gait.

Humanoid robots used in fashion shows typically rely on conventional locomotion policies optimized for balance, velocity tracking, and disturbance recovery. While these policies allow robots to traverse a runway, they do not address the expressive or garment-presentation functions of the catwalk. Prior work has examined expressive robot gait and isolated fashion-model-like motion, but there remains no general data-driven framework for learning multiple runway styles from human demonstrations and transferring them to physical humanoids. We therefore formulate robotic catwalking as a style-conditioned locomotion problem evaluated through both quantitative measures of motion and stability and qualitative judgments of style, confidence, and fashion-show appropriateness.

Video is a simple way to provide such a motion. Instead of manually creating robot trajectories, the user can show the desired behavior in a monocular video. The main difficulty is that motion recovery, retargeting, training, and deployment are usually handled by separate tools with different formats and configurations.

In this work, we present an end-to-end pipeline that takes a human video and produces a policy ready to run on a humanoid robot. The user provides the video and selects the target robot, while the pipeline handles motion recovery, retargeting, correction, training, simulation testing, and deployment.

We demonstrate the pipeline on the Booster K1 using runway-style catwalk motion. Catwalk walking is our main example, but the same workflow can be used for other custom motions in fashion, entertainment, and related applications.

%% file: sections/related_work.tex
\vspace{-1 mm}
\section{RELATED WORK}
\vspace{-1 mm}
Gravity-View Human Motion Recovery (GVHMR)~\cite{gvhmr} converts a video into 3D human motion aligned with the ground. General Motion Retargeting (GMR)~\cite{gmr} then adapts the motion to the selected robot.

BeyondMimic~\cite{beyondmimic} uses the retargeted motion to train a tracking policy in the Booster training framework~\cite{boostertrain}. The policy is tested and run on the robot with the Booster deployment framework~\cite{boosterdeploy}. These frameworks cover separate stages of the process but do not provide a single workflow from video to robot.

Previous work has explored fashion-inspired and expressive robot walking. Or~\cite{fashionmodelrobot} simulated fashion-model-like motions on a humanoid, while Huzaifa et al.~\cite{expressivegait} generated and evaluated different expressive gaits. Kobayashi et al.~\cite{catwalkstyles} identified leg crossing as a difference among runway styles, motivating our use of step width to compare the learned policy with the standard K1 walk.

%% file: sections/method.tex
\section{METHOD}

\begin{figure}[t]
    \centering
    \includegraphics[width=\columnwidth]{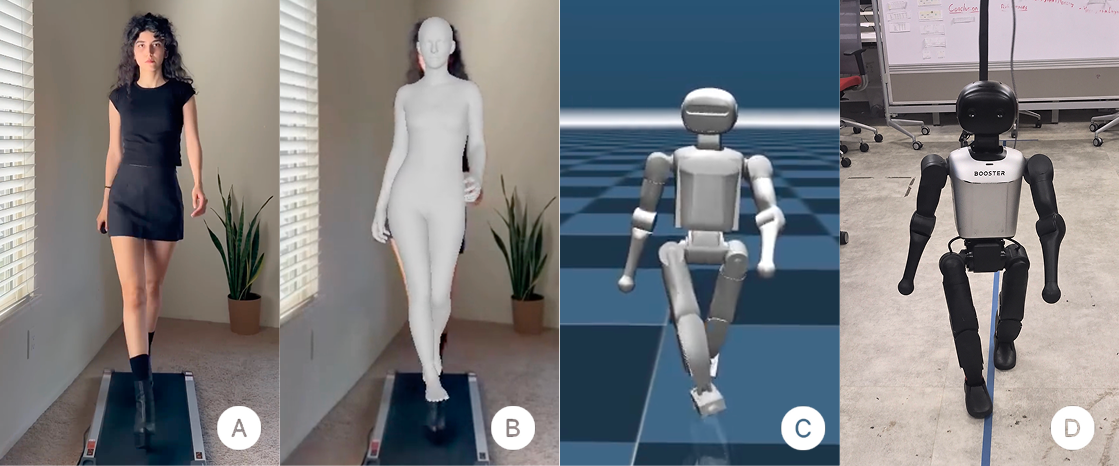}
    \caption{Overview of the pipeline: (a) monocular runway video, (b) recovered 3D human motion, (c) retargeted robot motion in simulation, and (d) physical deployment on the Booster K1.}
    \label{fig:pipeline_strip}
\end{figure}

Our pipeline turns a monocular video of human motion into a policy that can be tested and run on a humanoid robot. The user provides the video and selects a robot supported by GMR~\cite{gmr}. The pipeline handles 3D motion generation, retargeting, motion correction, policy training, and deployment. An overview of the stages is shown in Fig.~\ref{fig:pipeline_strip}.

\subsection{Motion Recovery}

The pipeline uses GVHMR~\cite{gvhmr} to convert the input video into 3D human motion. It reads the video's frame rate and selects the setting for a static or moving camera. The resulting SMPL-X motion and global path are then sent to the retargeting stage.

\subsection{Robot Retargeting}

The user selects a robot supported by GMR~\cite{gmr}. GMR adapts the human motion to the robot's body proportions and joints. The output is a motion trajectory for that robot.

\subsection{Motion Correction and Conversion}

After retargeting, the pipeline detects foot contacts and adjusts the motion to keep the feet aligned with the floor. It also removes vertical drift without changing the joint angles, preserving the original walking style. Finally, the motion is converted into the training format with the robot joint states, base pose, and frame timing from the source video.

\subsection{Policy Training}

The processed motion is used as the reference trajectory for training with BeyondMimic~\cite{beyondmimic} in the Booster training framework~\cite{boostertrain}. We use the existing training method while keeping the actuator limits, stiffness, damping, and action scaling consistent with the deployment configuration. This avoids using a policy on hardware with different control settings from those used during training.

\subsection{Simulation and Deployment}

After training, the exported policy is loaded into the Booster deployment framework~\cite{boosterdeploy} and tested in MuJoCo using the deployment controller. This sim-to-sim test is used to check balance, tracking, and joint behavior before hardware execution. The same exported policy can then be launched on the physical robot with our deployment script. The final output of the workflow is therefore a runnable robot policy rather than only a retargeted trajectory or training checkpoint.

%% file: sections/experimental_setup.tex
\section{EXPERIMENTAL SETUP}

The catwalk policy was trained using BeyondMimic in the Booster training framework and first tested in MuJoCo. The trained policy was then deployed on the Booster K1 through a custom deployment script and a wired Ethernet connection.

A straight 5 m walking path was marked on the floor using tape. The markings indicated the starting position, runway centerline,and distance intervals used to measure the traveled distance and step width. Each test contained approximately 23 steps. Safety belts were attached to the robot to prevent damage in the event of a fall while remaining loose during normal walking.

%% file: sections/results.tex
\section{RESULTS}

\subsection{Physical Execution}

The learned policy completed all 20 physical trials without falling, with approximately 23 steps in each trial. The robot showed narrow foot placement and coordinated leg, torso, and arm movements. The first step sometimes caused a small lateral offset, after which the robot continued along a stable path approximately parallel to the runway line.

\subsection{Foot-Placement Comparison}

We compared the lateral distance between consecutive foot placements using 12 steps from the standard K1 walk and 23 steps from the learned catwalk policy. Negative values indicate crossover steps, where one foot was placed beyond the other foot laterally.

The standard K1 walk produced step widths between 5.1 and 11.4 cm, while the catwalk policy stayed between $-0.8$ and 1.8 cm. The learned policy therefore achieved substantially narrower foot placement, with occasional crossover steps, compared with the standard K1 walk.

%% file: sections/future_work.tex
\section{FUTURE WORK}

Future work includes developing a controllable policy, creating a larger catwalk motion dataset, improving the evaluation metrics and involving fashion experts in assessment of the robot's walking style and training and deploying the pipeline on humanoid robots beyond the Booster K1.